\documentclass[arxiv,dvipsnames,format=sigconf,anonymous=false,review=false]{acmart} 
\usepackage{rotating,graphicx}
\usepackage{subcaption}
\setcopyright{none}
\renewcommand\footnotetextcopyrightpermission[1]

\usepackage{paralist}
\usepackage{url}
\usepackage{xcolor}
\usepackage{xspace}
\usepackage{graphicx}
\usepackage{url}
\usepackage[linesnumbered,ruled,noend]{algorithm2e}
\usepackage{bm}
\usepackage{booktabs}
\usepackage{array,multirow}
\usepackage{hyperref}

\newcommand{\cccFour}{CCC4\xspace}

\begin{document}

\pagestyle{plain}

\title[]{Evolutionary and Coevolutionary Multi-Agent Design Choices and Dynamics}   



\author{Erik Hemberg}
\email{hembergeri@csail.mit.edu}
\affiliation{MIT CSAIL
\country{US}
}

\author{Eric Liu}
\email{eliu3765@mit.edu}
\affiliation{MIT CSAIL
\country{US}
}

\author{Lucille Fuller}
\email{lucifull@mit.edu}
\affiliation{MIT CSAIL
\country{US}
}

\author{Stephen Moskal}
\email{smoskal@mit.edu}
\affiliation{MIT CSAIL
\country{US}
}

\author{Una-May O'Reilly}
\email{unamay@csail.mit.edu}
\affiliation{MIT CSAIL
\country{US}
}

\begin{abstract}
We investigate two representation alternatives for the controllers of teams of cyber agents. 
We combine these controller representations with different evolutionary algorithms, one of which introduces a novel LLM-supported mutation operator. 
Using a cyber security scenario, we evaluate agent learning when one side is trained to compete against a side that does not evolve and when two sides coevolve with each other.  
This allows us to quantify the relative merits and tradeoffs of representation and algorithm combinations in terms of team performance. 
Our versions of grammatical evolution algorithms using grammars that allow a controller to be expressed in code-like logic can achieve the best team performance.
The scenario also allows us to compare the performance impact and dynamics of coevolution versus evolution under different combinations. 
Across the algorithms and representations, we observe that coevolution reduces the performance highs and lows of both sides while it induces fluctuations on both sides.  In contrast,  when only one-side is optimized, performance peaks are higher  and is more sustained than when both sides are optimized with coevolution. 

\end{abstract}
\maketitle

\section{Introduction}


With larger scale and closer-to-reality scenarios, complex challenges aim to push the limits of autonomous agent training.
The CybORG gym, an engine that models cyber networks, threats, and security response,  has supported several such challenges~\cite{Standen2021CybORGAG}.
A recent example is the Cyber Cage Challenge 4 (\cccFour).
It expects cyber agents to be designed by means of machine learning.
It  presents a fictitious scenario (with stochastic variants) of two competing nations. One nation (red) seeks to disrupt the server  which handles the other's (blue's) mission. The server is within a complex cyber environment with  four networks, five security zones defined over the networks, and various firewall defenses.
In the scenario there are multiple "blue" agents comprising a team that are positioned in different zones. Blue must defend the environment by taking actions against a multi-phase attack controlled by  red.

The \cccFour scenario poses the fundamental problem of learning a sequential decision controller  for an agent or team of agents. The agent's controller, with only partial observability must  decide  two things, one simulation step at a time:   the  host on which it will target its next action, and what that action will be.  Additionally, the multiple agents must work in coordination while being  positioned in different zones. Their actions realistically model the defender's preventative, monitoring and repairing options.  CybORG provides easy performance comparison among controllers and teams with a point-based system of rewards for different accomplishments.

While \cccFour requires multi-agent training, this training  is confined to blue teams.   Because we are interested in training autonomous red and blue cyber agents, we extend \cccFour in two ways. Red teams can also be trained and, second, more information about the scenario is made accessible through new observation functions. 

This leads to the question of how to best train red and blue teams when they compete against each other.   To start, we consider controller design in terms of representation and algorithm choices. These  are critical to agent training and agent performance.  A designer must decide what part of the solution, in this case the agent's controller,  to hard code versus what part they ask the Evolutionary Algorithm~(EA) to optimize for them. They can opt for largely predetermined solution logic that sets up a smaller and simpler search space for optimization. Or they can hard code only basic solution logic and  provide the EA with  powerful logic elements that must be composed into  a controller. This implies the EA must optimize in a more complex and larger  search space.  Choosing between these two extremes or finding a design choice between them is an open question in the challenge setting we have created. 
To address this, we explore  agent controller representations and their variants which have different search space size and expressivity tradeoffs.  Our research question is: What is the best combination of algorithm and representation that achieves the best performing red and blue teams? 

Our interest in competing cyber agents that take offensive and defensive sides motivates a related interest in adversarial dynamics. 
Defenses in cyber systems are mostly optimized to repel specific attackers whose activities are well characterized, see, for example, the \cccFour Challenge itself.  We seek optimizations that assume attackers will evolve and adapt counter measures to defensive measures. 
Though two-sided competition and optimization are more challenging in complexity, modeling them is important to better understand the dynamics~\cite{popovici2012,krawiec2016solving}. 
Our second research question centers on this model of adversarial dynamics.
How do team performance levels change over time as the two sides adapt while competing?  
Does peak performance of either side diminish when both sides co-optimize versus when they are optimized independently?   
Can we expect to see more or less rewards going to each team?   
To answer this second set of research questions, our baseline is when one side (red or blue) evolves and optimizes with combinations of representations and evolutionary algorithms, while the other (blue or red respectively) does not. We then compare baseline performance with the performance of our combinations of representations and coevolutionary algorithms. 


 Our contributions follow:
 
Around design choices, one of our two representations is an action
selection matrix, indexed by state and action.  For each state, the
likelihood of each action needs to be optimized. This probability can
be optimized by a genetic algorithm~(GA)~\cite{Goldberg1989} or
evolutionary strategy~(ES)~\cite{back1996evolutionary}.  Alternatively,
we use grammars that are optimized by grammatical evolution~(GE)\cite{o2001grammatical}.  We explore progressively more powerful
grammars which sets up progressively larger search spaces.  Each of
these algorithms, GA, ES and GE, is also integrated into their own
coevolutionary algorithm variant.


\begin{itemize}
\item We compare designs with the same controller being referenced by
  all members of a team versus designs where each team member
  references its own controller when the grammar variants are used. We
  find the use of one controller per team can sometimes perform as well as multiple controllers per team.
\item We introduce a version of GE that incorporates a large language
  model as part of its mutation operator. We compare this algorithm to
  GE in both one-side learning (with 1 GE algorithms) and two-sided
  learning (2 GE algorithms connected by competition) and conduct a
  small study into the LLM supported operator. We observe that
  the LLM operator is very sensitive to the LLM-model and prompt used. 
\item Given the action selection matrix representation, we compare the
  performance under continuous or discrete parameter optimization. We
  find that the discrete parameter optimization is a better choice
  since it works for both cases and is a better choice for red.
\item We compare the grammar variants using the same algorithm, asking
  how their expressiveness impacts team performance. We find that the
  grammar expressiveness can influence the performance and that a larger
  grammar can perform well with the correct terminals.

\end{itemize}

Around adversarial dynamics we run four comparisons. In each we compare the training dynamics of a population's mean performing and best performing controllers under one-sided learning and under coevolution for four algorithm and representation combinations. We also look for properties of the dynamics that differ or remain consistent across these comparisons. We find that coevolution consistently blunts the performance highs and lows of both sides while it induces fluctuations in both sides.  In contrast,  when only one-side is optimized, performance peaks higher  and is more sustained than when both sides are optimized with coevolution. 


The paper proceeds as follows. In Section~\ref{sec:prelims} we present
background. In Section~\ref{sec:related_work} we present related
work. In Section~\ref{sec:method} we present our methodology. In
Section~\ref{sec:experiments} we present experiments. In
Section~\ref{sec:conclusions} we present conclusions and future work.
  
\section{Background}\label{sec:prelims}


\subsection{CybORG}

\begin{figure*}[!h]
  \centering
    \captionsetup{font=footnotesize}
  \includegraphics[width=0.992\textwidth]{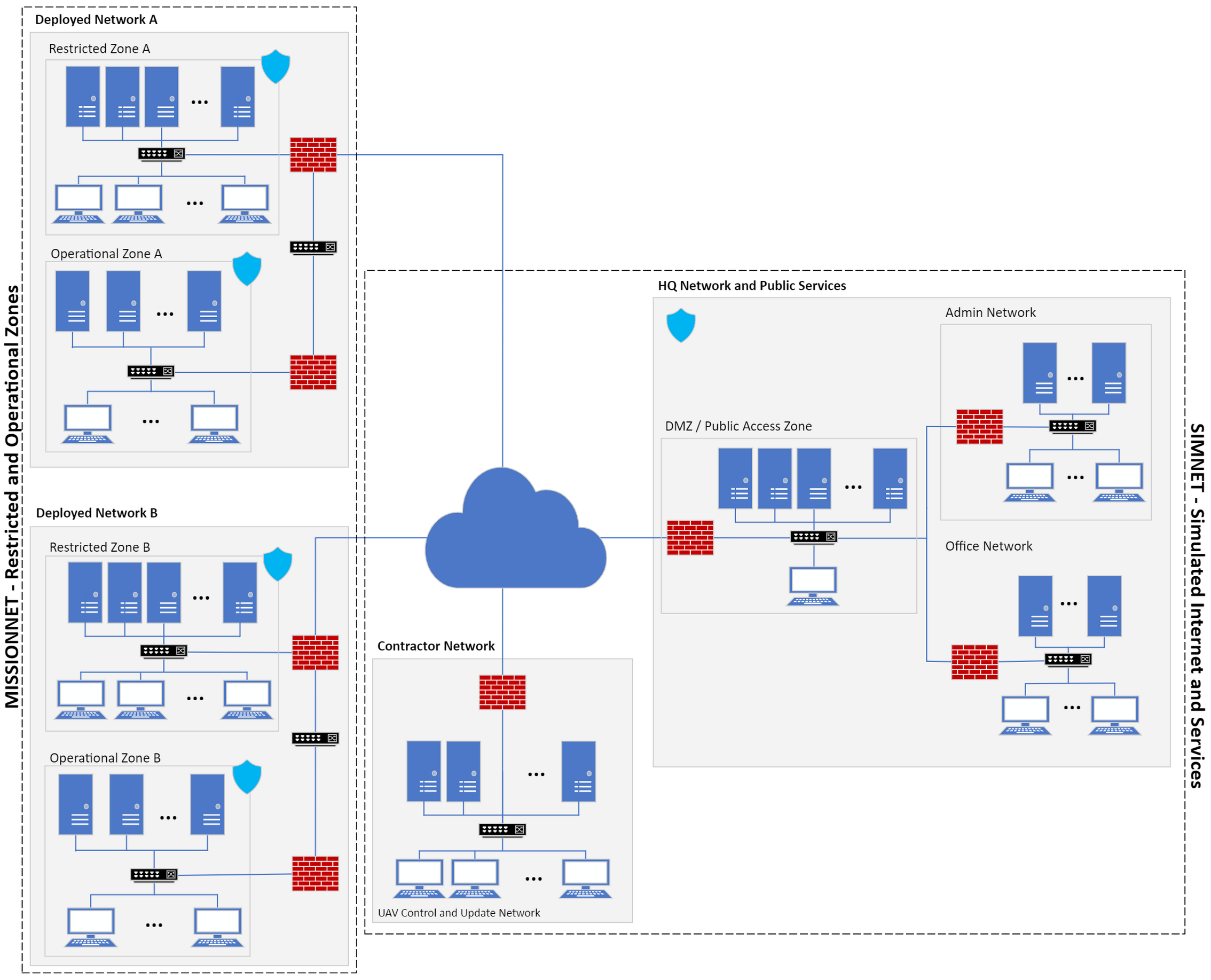}
  \caption{CybORG Cage Challenge 4 network.}
  \label{fig:ccc4}
\end{figure*}

We use the CyBORG gym~\cite{Standen2021CybORGAG}. For more details,
including the mission phases, see Appendix~\ref{appendix:CybORG}.
The \cccFour network consists of four subnetworks, see
Figure~\ref{fig:ccc4}. Two are deployed~(A, B), one is the
Headquarters (HQ) and another is the Contractor. These networks
connect via the internet. Each deployed network has two
security zones: a restricted zone and an operational zone. The HQ
network has three security zones: a Public Access Zone, an Admin Zone
and an Office Network. The Contractor network only contains a single
UAV control zone~\cite{CCC4,cage_cyborg_2023}.

The number of hosts in each security zone and their services are
randomized. Each zone has 1-6 servers and 3-10 user hosts. Each host
and server have 1-5 services. The network has 5 blue agents. Each
deployed network has one blue agent in each security zone. The HQ has
one blue agent for all zones. The Contractor network is undefended. On
every host a user is represented by green agents. Green agents
randomly do local work or randomly reach out to external services,
either in their local or a remote zone.

Red begins with access to a random machine in the contractor
network. Every turn there is a low probability that a red agent spawns
if a green agent opens a phishing email. Red can also spawn in
networks, when a green user accesses a compromised service. There can
only exist one red agent in each zone, though they can maintain a
presence on multiple hosts. Red always maintain a foothold in the
Contractor Network.

\textit{Action sets} for agent actions have a specified duration, which varies
depending on the action chosen. Agents must wait until their action is
completed before they perform another action. All agents can take the
action Sleep.

Red actions are:
\begin{inparaenum}
\item Discover Remote Systems
\item Aggressive Service Discovery
\item Stealthy Service Discovery
\item Exploit Network Services
\item Privilege Escalate
\item Impact
\item Degrade Services
\item Discover Deception
\end{inparaenum}

Blue actions are:
\begin{inparaenum}
\item Monitor
\item Analyze
\item Deploy Decoy
\item Remove
\item Restore
\item Block Traffic
\item Allow Traffic  
\end{inparaenum}

Green actions are:
\begin{inparaenum}
\item Access Service
\item Local Work
\end{inparaenum}

\textit{Observations} are events from device configurations~(files and
users) and processes. Blue observes events on all devices. Red only
observes events on discovered devices. \textit{Rewards} are tied to
green agent actions, see Appendix~\ref{sec:cyborg-cage-chall}. Blue
receives penalties~(and red rewards) when a green agent is unable to
perform an action, regardless of if it is a red or blue action preventing
it. Green agents in mission-critical zones generate higher penalties
when their mission is active.

\subsection{Algorithm Choices}

Biological coevolution refers to the influences two or more species
exert on each other's evolution~\cite{rosin1997new}.  A seminal paper
on reciprocal relationships between insects and plants coined
``coevolution''~\cite{ehrlich1964butterflies}.  Coevolution can be
cooperative, i.e., mutual benefit, or competitive, i.e., interactions
arising from contested resources.  An EA typically evolves individual
solutions, e.g., fixed length bit strings~\cite{Goldberg1989} using a
predefined defined fitness function to evaluate quality. In contrast,
coevolutionary algorithms~(CA) calculate an individual's fitness based
on its interactions with other individuals or a dynamic environment
allowing them to mimic coupled natural species-to-species
interactions. This contribution extends the prior work studying
coevolutionary
algorithms~\cite{Antonio2018,popovici2012,rosin1997new,sims1994evolving,mitchell2006coevolutionary,krawiec2016solving,hemberg2016detecting,o2020adversarial}.

The dynamics of CAs are difficult to analyze because their fitness
evaluation is derived \textit{interactively},~\cite{popovici2012}.
Effectively an individual's fitness is a sample-based estimate of their
performance where the samples are drawn from the other population
which itself is evolving. A more formal approach, using the
\textit{solution} and \textit{test} perspective, describes fitness
assignments as \textit{solution
  concepts}~\cite{popovici2012}.

\section{Related Work}
\label{sec:related_work}

Table~\ref{tab:rw_overview} summarizes work related to the agent training in a number of cyber security-oriented gyms, including CybORG, and CyberBattleSim. 
Closely related to this work is ~\cite{Heckel2023NeuroevolutionFA} where neuroevolution was used for to
evolve Deep Reinforcement Learning agents in a CybORG environment.
As well, competitive coevolutionary agent-learning for
cybersecurity  was investigated in the CyberBattleSim
environment~\cite{Shashkov2023AdversarialAF}. Here the investigation
was a comparison of different algorithms that used the same agent representation.
Finally, early work on a coevolutionary agent-based network defense used a lightweight event system
(CANDLES)~\cite{Rush2015CoevolutionaryAN} represented attack and defense agents with rules.

 We note that there exists a gap in the related work regarding studying adversarial 
dynamics and exploring different agent representations and evolutionary
computation methods.

\begin{table*}[htb]
  \centering
  \footnotesize
  \caption{Overview of work at intersection of learning agents with
    coevolutionary algorithms. Columns show the agent environment~(Env.),
    agent representation~(A. repr.), evolutionary computation method~(EC), and
    learning dynamics~(Dyn.).}
  \label{tab:rw_overview}
  \begin{tabular}{l|p{10cm}llll}
    \textbf{Ath.} & \textbf{Title} & \textbf{Env.} & \textbf{Agent Repr.} & \textbf{EC?} & \textbf{Coev?}\\
    \hline
    \multicolumn{6}{c}{\textit{Cyber Security Environments}} \\
    \hline
    Us & Agent representation and heuristics in CybORG & CybORG CAGE-4 & FSM, Rules & Y & Y \\
    \cite{hammar2024optimal} & Optimal Defender Strategies for CAGE-2 using Causal Modeling and Tree Search & CybORG CAGE-2 & Structual Causal Model & N & N \\
    \cite{Heckel2023NeuroevolutionFA} & Neuroevolution for Autonomous Cyber Defense & CybORG & DRL & Y & N \\
    \cite{Shashkov2023AdversarialAF} & Adversarial agent-learning for cybersecurity: a comparison of algorithms & CyberBattleSim & DRL & Y & Y \\
    \cite{Rush2015CoevolutionaryAN} & Coevolutionary Agent-based Network Defense Lightweight Event System & CANDLES & Rules & Y & Y \\
    \cite{Smith2016InitiatingAM} & Initiating a Moving Target Network Defense with a Real-time Neuro-evolutionary Detector & DNN & NEAT & Y & N \\
     \cite{Maliukov2024EvolvingAC} & Evolving Assembly Code in an Adversarial Environment & Code Guru & Assembly & Y & Y \\
    \cite{Wei2024OfflineRL} & Offline Reinforcement Learning for Autonomous Cyber Defense Agents & CyberVAN & Rules & N & N \\
       \hline
    \multicolumn{6}{c}{\textit{Theoretical Studies}} \\
      \hline
     \cite{goel2024hardening} & Hardening Active Directory Graphs via Evolutionary Diversity Optimization based Policies & Stackleberg Game & DRL & Y & Y \\
   \cite{Fajardo2024ASC} & A Self-adaptive Coevolutionary Algorithm & Defend-It & Discrete & Y & Y \\
   \hline
    \multicolumn{6}{c}{\textit{Other Environments}} \\
      \hline
    \cite{Harris2021CompetitiveCF} & Competitive coevolution for defense and security: Elo-based similar-strength opponent sampling & Predator-Prey & Y & Y & Y \\
    \cite{Kelly2017EmergentTG} & Emergent Tangled Graph Representations for Atari Game Playing Agents & Atari Games & Code & Y & N \\
  \end{tabular}
\end{table*}

\section{Representation Choices and Algorithms}
\label{sec:method}

\subsection{\textbf{Representation~1}: Action Selection Matrix}
An agent controller is a matrix of $n$ rows, one per state and $m$
columns, one per action.  Each cell$(i,j )$ encodes the probability of
choosing an action $j$ in state~$i$.

\textit{Optimization of Continuous Probabilities with an ES:} When the
representation of each action probability is a floating point, the
probabilities are optimized by an ES~\cite{back1996evolutionary}.  Our
ES implementation uses uniform random initialization of solutions. For variation, we use Gaussian mutation.

\textit{Optimization of Discretized Probabilities with a GA:} Per
\cccFour the representation of each action probability is discretized
to four choices: $(0.0, 0.25. 0.5, 0.75)$.  and a Genetic Algorithm
which uses integer vectors~\cite{Goldberg1989} is used for
optimization.  Our GA implementation uses uniform random
initialization of solutions. For variation, we use random integer
mutation and one point crossover.  Evolved values are normalized
across all actions for each state.

\subsection{\textbf{Representation~2}: Context Free Grammars that Express a Block of Python Code}
In this representation there are no agent states. Instead, a
controller is some Python code block~(in some cases a function),
structured to be executable in the CybORG environment.  A context free
grammar $ G \in \mathcal{G}$ expresses the controller code.  It
includes elements that do not evolve but that are necessary for
syntax, such as the block's function header, return values and
reserved words such as ``\texttt{if}''.  The start symbol is
\texttt{statements} and the terminals are actions, targets, and
constants.  Non-terminals include the test and true branch elements of
an \texttt{if}-statement and the observation functions that require an
observation argument and return a count of the observed item.  See
Figure~\ref{fig:grammarBaseline} for part of the baseline
grammar~(full is in Appendix~\ref{sec:code-grammars}).

\begin{figure}
\begin{small}
\begin{verbatim}
sections: "def action_and_target_selection(observation, name):
            "#Select action"
            statements
            "return action, target_heuristic"          
statements: statement 
          | statement
            statements
statement: "if" conditions ":"
               statement
         |  "action =" actions
conditions: operator "and" operator
          | operator "or" operator  | operator
operator: observations operand constant
         | success "== observation['success']"
operand: ">"    | "<"   | "=="
success: "TRUE"  | "FALSE"  | "UNKNOWN"
observations: "connections(observation)"
            | "files_user(observation)"
            | "files_root(observation)"
            | "n_servers(observation)" 
            | "root_access_levels(observation, name)"
constant: "0"  | "1"  | "2"
actions: "DiscoverRemoteSystems" 
       | "AggressiveServiceDiscovery" 
       ...
\end{verbatim}
\end{small}
\caption{Partial baseline grammar for a controller.}
\label{fig:grammarBaseline}
\end{figure}


The baseline grammar has three target selection options and two
observations function options.  Five variants of this baseline explore
changes in search space size and what is included in the
grammar. Three of the five variants remove the ability to select a
target from a set of three choices (newest, random, or oldest host on
agent's host list). Each hardcodes a different choice. The fourth
variant adds conditional statements to the baseline. While this
increases the search space, it provides more nuanced controller
designs.  The fifth variant adds three observations to the original
set of two. This gives the controller more discriminatory power to
base actions or target selection on knowledge returned from
observations.

\subsubsection{Algorithms for Grammars}

An integer vector can guide the generation (rewriting) of grammar
productions.  We use Grammatical Evolution (GE) to optimize this
vector~\cite{o2001grammatical,custode2024comparing}.  Our GE implementation uses uniform
random initialization. For variation, we use random integer mutation
and one point crossover. The decoding of the population uses a context
free grammar. If the decoding is invalid a new random individual is
generated.

Our GE-LLM implementation uses uniform random initialization of
solutions with a grammar as in GE. For variation, however we use an
LLM-supported mutation. The mutation operator works on the code block,
not the genome. The code block and other contextual information and a
mutation task description are sent to a LLM. The LLM is expected to
return a mutated block of code. The underlying question is whether the
LLM's proficiency at code generation and possible training experience
with domain-related information can contribute to semantically
interesting mutations.  LLM based operators of this type are often
problem environment dependent for this
reason~\cite{hemberg2024evolving}. Note this is another contrast with
canonical EA based operators that are problem agnostic, in addition to
not working on the genome.

\begin{figure}
  \centering
    \captionsetup{font=footnotesize}
  \includegraphics[width=0.495\textwidth]{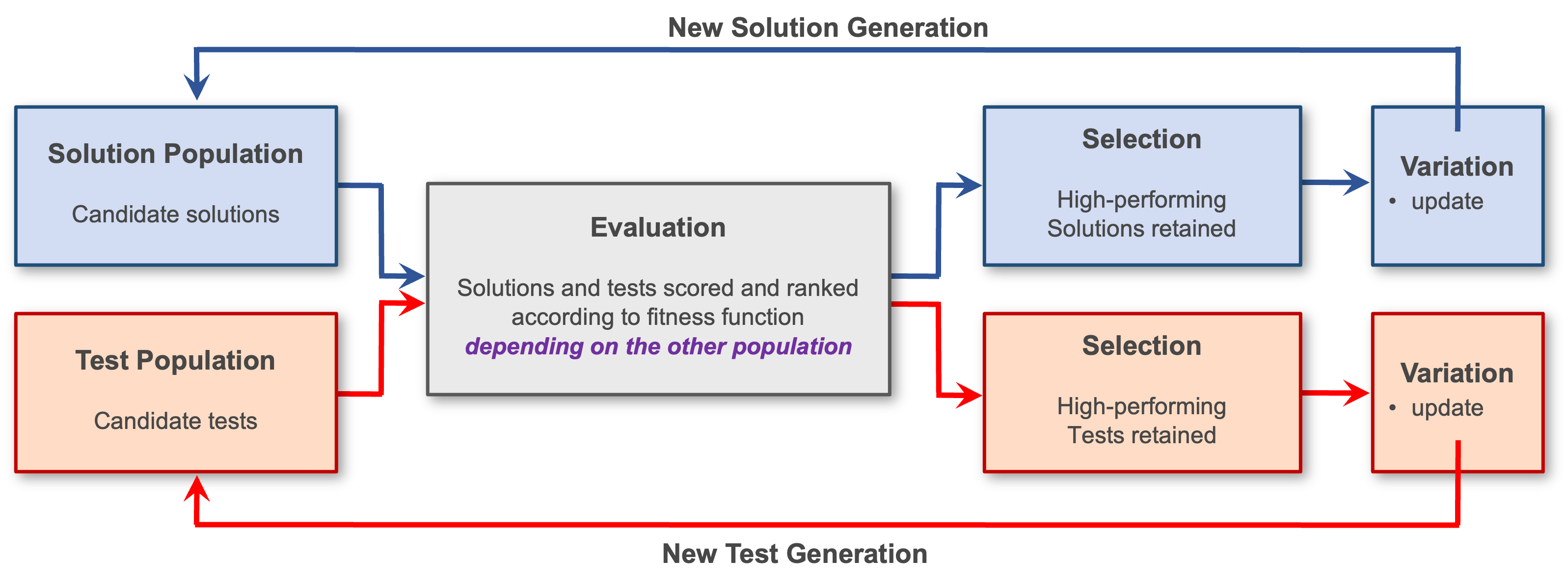}
  \caption{A coevolutionary algorithm unites two EAs, here called Solution and Test, coupling them at the fitness evaluation step.}
  \label{fig:coevAlg}
\end{figure}

A Competitive Coevolutionary algorithm~(CCA) has two competing
populations with conflicting objectives engaged in a mini-max search.
One way to describe a CCA is to note that it is really two EAs that
run independently except for a coupling at their fitness function
evaluation steps. At that step, one member from each EA's population
is selected and competed against a member from the other EA's
population. The result of the competition (and any observations
arising from the competition) is sent to each EA where it is
transformed into a fitness score. See Figure~\ref{fig:coevAlg} for an
illustration.  Our coevolutionary algorithms are couplings of two
GAs~(GA-C), two ES algorithms~(ES-C), and two GE and GE-LLM algorithms
(GE-C, and GE-LLM-C) yield three basic variants of a CCA and two
variants of a coevolutionary GE algorithm.  These are labeled in our
plots as GA-C, ES-C, GE-C, and GE LLM-C.

\section{Experiments}
\label{sec:experiments}

In Sec.~\ref{sec:setup} we describe the experimental setup and
design. In Sec.~\ref{sec:results--analysis} we report the results. In
Sec.~\ref{sec:discussion} we present a discussion.

\subsection{Setup}
\label{sec:setup}

In \cccFour we add two functions that extract from the red agent's
observations dictionary. They are respectively the number of hosts it
has discovered, and the number of sessions in which it has root
access. These are updated after every action.  We support red agent
selection of where their action will execute by maintaining a list of
known hosts, ordered by when they are discovered. The host acquired by
the oldest or newest access can be selected or the selection can be
random. We also add a blue agent to act as an (non-learning) adversary
to red. It uses a finite state machine to move from state to state,
using the provided set of blue actions, conditioned on success of an
action, see Appendix~\ref{sec:finite-state-mach}.

The experimental settings are shown in
Table~\ref{tab:experimental_setup}. Due to the cost of running the
CybORG simulation we use a limited number of steps and fitness
evaluations. Note, we use all-vs-all evaluation for coevoltion which
increases the number of fitness evaluation per iteration by the square
of the population. We chose a quite high mutation rate since we are
interested in exploring the search space.

\begin{table}[h]
  \centering
  \small           
    \caption{Experimental setup.}
  \label{tab:experimental_setup}
  \begin{tabular}{l|l}
    \textbf{Name} & \textbf{Value}\\
    \hline
    \multicolumn{2}{c}{\textbf{CybORG}} \\
    \hline
    Steps & 75 \\
    Repetitions & 2 \\
    Red team agents & 6 \\
    Blue team agents & 5 \\
    Grammar & See Appendix~\ref{sec:code-grammars} \\
    Finite State Machine & See Appendix~\ref{sec:finite-state-mach} \\
    Fixed red agent & FSM \\
    Fixed blue agent & FSM \\
    \hline
    \multicolumn{2}{c}{\textbf{Evolutionary Algorithm}} \\
    \hline
    Trials & 6 \\
    Population Size & 10 \\
    Elite size & 1 \\
    Tournament Size & 2 \\
    Iterations~(Generations) &  $\geq 20$ \\
    Crossover probability & 0.5 \\
    Mutation probability & 0.5 \\
    Red team length & 180 FSM, 1000 GE \\
    Blue team length & 180 FSM, 1000 GE\\
    Coevolution solution concept & Mean Expected Utility \\
  \end{tabular}
  \end{table}

\subsubsection{Experimental Design}

We investigate the agent representation, solution representation,
algorithm and adversary learning combinations shown in
Table~\ref{tab:experimental_design}. 
\begin{table}[h]
  \centering
  \small             
  \caption{Experimental design. A. Repr. is agent
    representation. S. Repr. is the solution representation. Alg. is
    the algorithm. Adv. is the fixed adversary, where coev means both
    are learning.}
  \label{tab:experimental_design}
  \begin{tabular}{l|llll}
    \textbf{Name} & \textbf{A. Repr.} & \textbf{S. Repr.} & \textbf{Alg.} & \textbf{Adv.}\\
    \hline
    \multicolumn{5}{c}{\textbf{Static Adversary}} \\
    \hline
    ES-B & FSM & Cont. & ES & Red \\
    ES-R & FSM & Cont. & ES & Blue \\
    GA-B & FSM & Discrete & GA & Red\\
    GA-R & FSM & Discrete & GA & Blue\\
    GE-B & Code & Grammar  & GE & Red\\
    GE-R & Code & Grammar  & GE & Blue\\
    GE-LLM-B & Code & Grammar & GE-LLM & Red\\
    GE-LLM-R & Code & Grammar & GE-LLM & Blue\\
    \hline
    \multicolumn{5}{c}{\textbf{Dynamic Adversary}} \\
    \hline
    ES-C & FSM & Cont. & ES & Coev \\
    GA-C & GA & Discrete & GA & Coev\\
    GE-C & Code & Grammar  & GE & Coev\\
    GE-LLM-C & Code & Grammar & GE, LLM & Coev\\
  \end{tabular}
  \end{table}

We investigate impact of controllers per team, grammars in the form of
target selection and extending the observations, as well as
LLM. Table~\ref{tab:experimental_design_detail} show the number of
controllers and grammar complexity designs.
\begin{asparadesc}
\item [One Controller] Only one controller is used for all the re/blue
  agents in the team. This results in a smaller search space, but less
  ability to refine behavior due to ``circumstances''
\item [Grammar complexity] Target is selected based on the
  order of event observations. It can be: random, oldest observation,
  newest observation or conditional statement to determine how to
  pick. With the conditional statement the search space becomes
  larger, since it is a recursive grammar

  Extend observation Functions to: count number of
  connection to a device, count number of user files on a device,
  count number files root files on a device. The added observation
  functions also increase the search space, but can provide salient
  information for the agent controller
\item [LLM] We test multiple LLMs of different makes and sizes :
  GPT-3.5, Phi-4
\end{asparadesc}

\begin{table}[h]
  \centering
  \small             
  \caption{Variations of Agent representation for controllers per
    team, and grammar complexity for target selection, and extending
    the observations.}
  \label{tab:experimental_design_detail}
  \begin{tabular}{l|l}
    \textbf{Name} & \textbf{Variation} \\
    \hline
    \multicolumn{2}{c}{\textbf{Number of Controllers per team}} \\
    \hline
    ES & One or Many \\
    GE & One or Many \\
    \hline
    \multicolumn{2}{c}{\textbf{Grammar complexity}} \\
    \hline
    GE-TR & Target Random \\
    GE-TN & Target Newest \\
    GE-TO & Target Oldest \\
    GE-TC & Target Conditional \\
    GE-OE & Extended Observation\\
  \end{tabular}
  \end{table}

\subsection{Results \& Analysis}
\label{sec:results--analysis}

We report the best and the mean fitness of the population for each
iteration over multiple trials. Note, the best fitness for coevolving
agents is relative to the adversaries that it was evaluated
against. Thus, even though we keep the best solutions between
iterations the fitness of the best individual can still decrease.
Note, for visual clarity we do not display the fitness variance among
the independent trials. In addition, the best reward for the blue team
is 0, and best for the red team is scenario dependent~(depending on the rewards and number of steps in simulation). Finally, we
observe that the fitness~(reward) from the dynamically evolved
solutions are often lower than the statically evolved for all out
experiments. 

\subsubsection{Adversarial Dynamics}
\label{sec:evol-dynam}

\begin{figure*}[!h]
  \centering
    \captionsetup{font=footnotesize}
\begin{subfigure}[b]{0.49\textwidth}
  \centering
  \includegraphics[width=\textwidth]{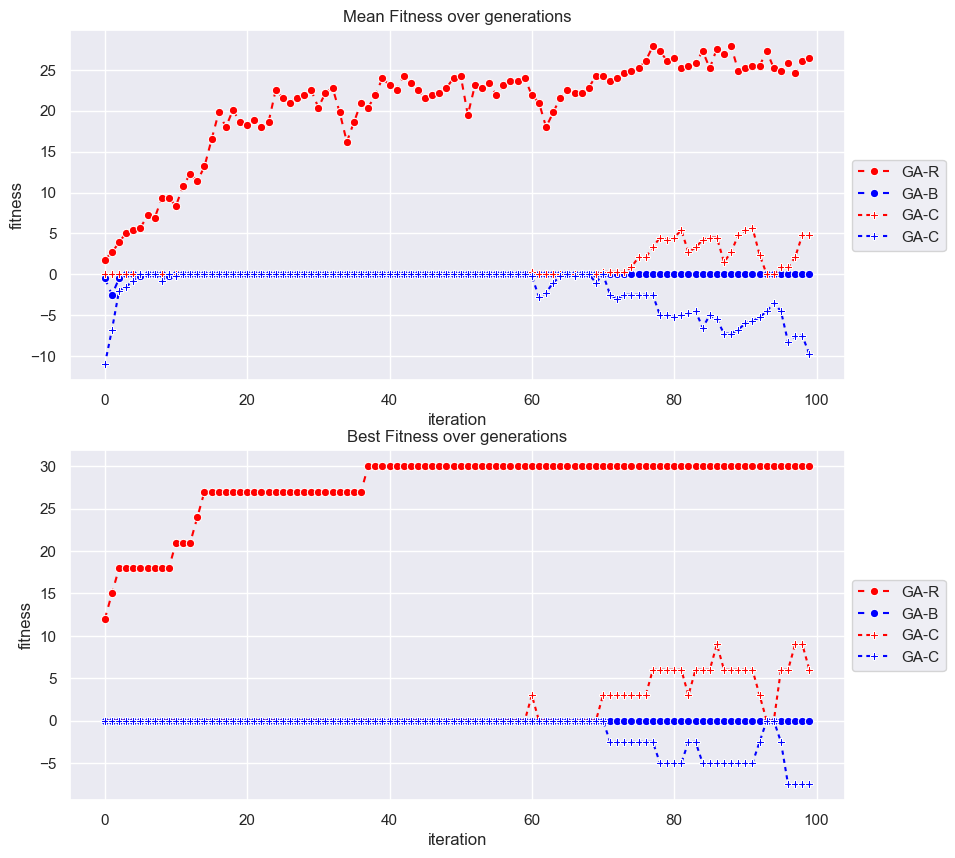}
  \caption{GA (FSM agent)}
  \label{fig:fsm_ga_c_results}
\end{subfigure}
\begin{subfigure}[b]{0.49\textwidth}
  \centering
  \includegraphics[width=\textwidth]{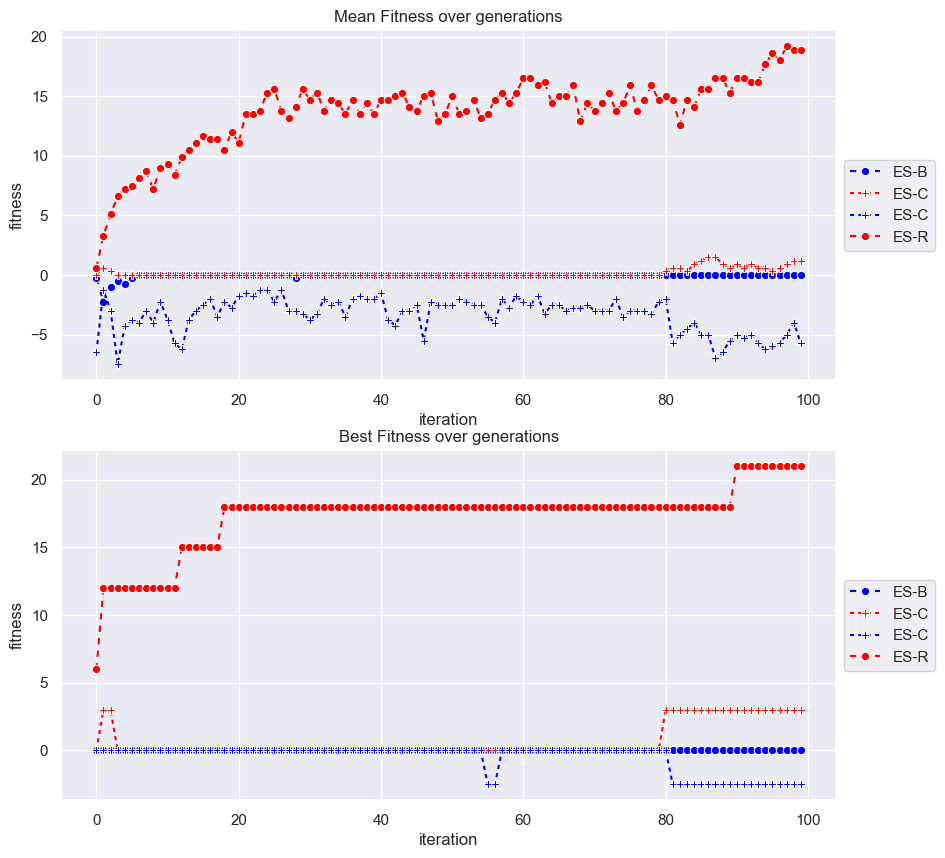}
  \caption{ES (FSM agent)}
  \label{fig:fsm_es_c_results}
\end{subfigure}

\begin{subfigure}[b]{0.49\textwidth}
  \centering
  \includegraphics[width=\textwidth]{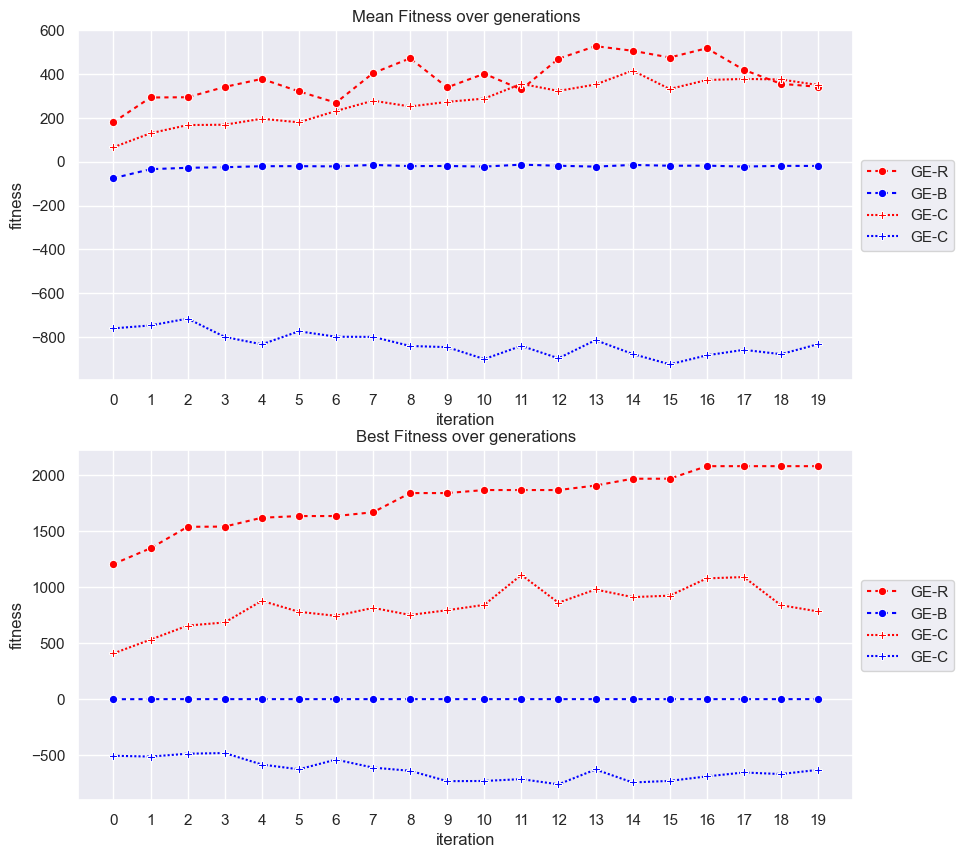}
  \caption{GE (Grammar agent)}
  \label{fig:ge_c_results}
\end{subfigure}
\begin{subfigure}[b]{0.49\textwidth}
  \centering
  \includegraphics[width=\textwidth]{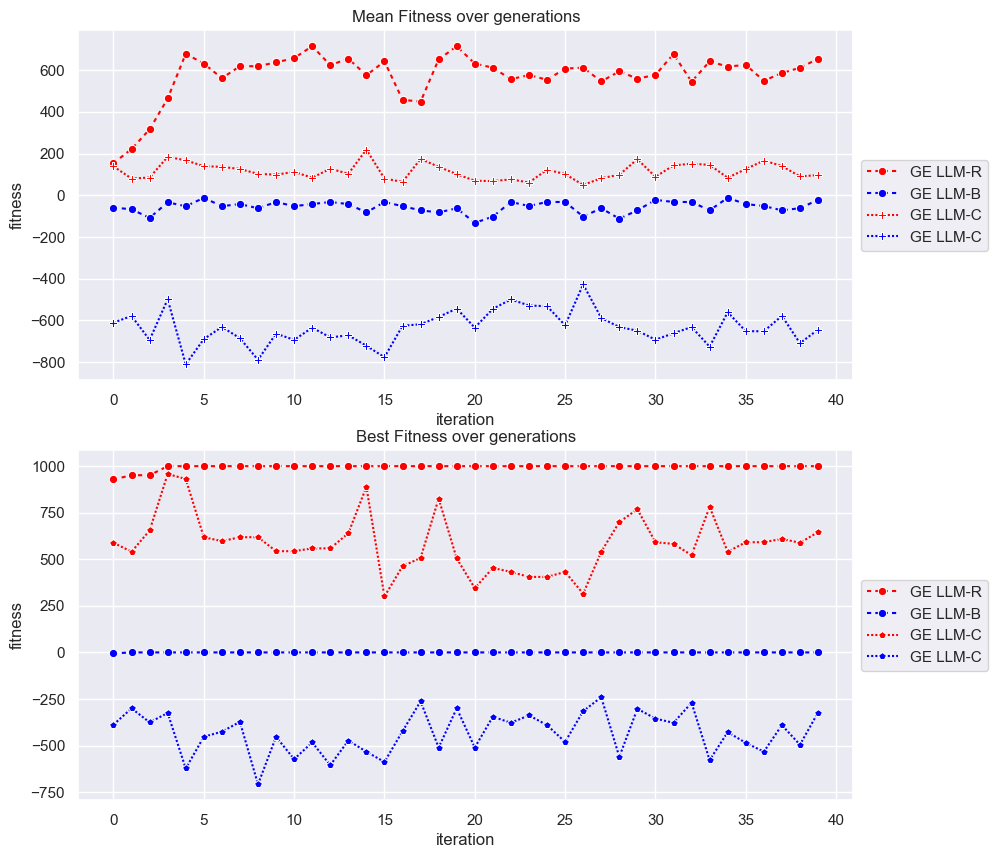}
  \caption{GE-LLM (Grammar agent)}
  \label{fig:ge_llm_c_results}
\end{subfigure}

\caption{Agent evolution in \cccFour. X-axis it the iteration. Y-axis
  agent fitness~(reward). One subplot shows best fitness the other
  mean population fitness. Lines show one-sided or two-sided
  evolution.}
  \label{fig:evolutionary_dynamics}
\end{figure*}

In Figure~\ref{fig:evolutionary_dynamics} we show the evolution of the
rewards for different algorithms~(and agent representations). We
observe that coevolution consistently dampens the reward highs and
lows of both sides while it induces fluctuations in both sides.  In
contrast, when only one-side is optimized, reward peaks higher and is
more sustained than when both sides are optimized with
coevolution. Note that the fitness GE are order of magnitude
higher than the FSM. This could indicate a limit in the FSM
representation and a benefit of the code rule representation.

The results for FSM agent evolution is shown in
Fig.~\ref{fig:fsm_ga_c_results} and \ref{fig:fsm_es_c_results}. Here
we observe fitness differences between a continuous~(ES) and a
discretized~(GA) FSM representation, the fitness can be higher for GA
than ES. We also see that when evolving blue agents and, for ES-B and
GA-B the randomly initialized blue agents always find a highest
performance, fitness 0. The mean in the population also converges to
the highest fitness. In contrast, for red agents for ES-R and GA-R the
red agents keep improving their fitness, with ES-R lower than GA-R.
For the coevolution of red and blue agents the ES-C have lower fitness,
and the blue agent rarely has 0 fitness. In contrast the GA-C plateaus
for several iterations.

The results for grammar agent evolution are shown in
Fig.~\ref{fig:ge_c_results} and \ref{fig:ge_llm_c_results}. We observe
that when evolving blue agents, for GE-B and GE-LLM-B the randomly
initialized blue agents always find a high performing agent, fitness
0. However, the mean in the population does not converges to the
highest fitness. The fitness variance is higher for the GE-LLM due to
the LLM variation creating more invalid solutions. For evolving red
agents, for GE-R and GE-LLM-R the red agents keep improving their
fitness, with GE-LLM-R lower than GE-R.  For the coevolution of red
and blue agents, again the red and blue have lower fitness. In
addition, the GE-LLM-C has lower fitness for both red and blue, still
order of magnitude lower than FSM. For GE-C there are indications of
oscillation of the red and blue fitness.

\subsubsection{Number of Controllers per team}

\begin{figure*}[!h]
  \centering
    \captionsetup{font=footnotesize}
\begin{subfigure}[b]{0.49\textwidth}
  \centering
  \includegraphics[width=\textwidth]{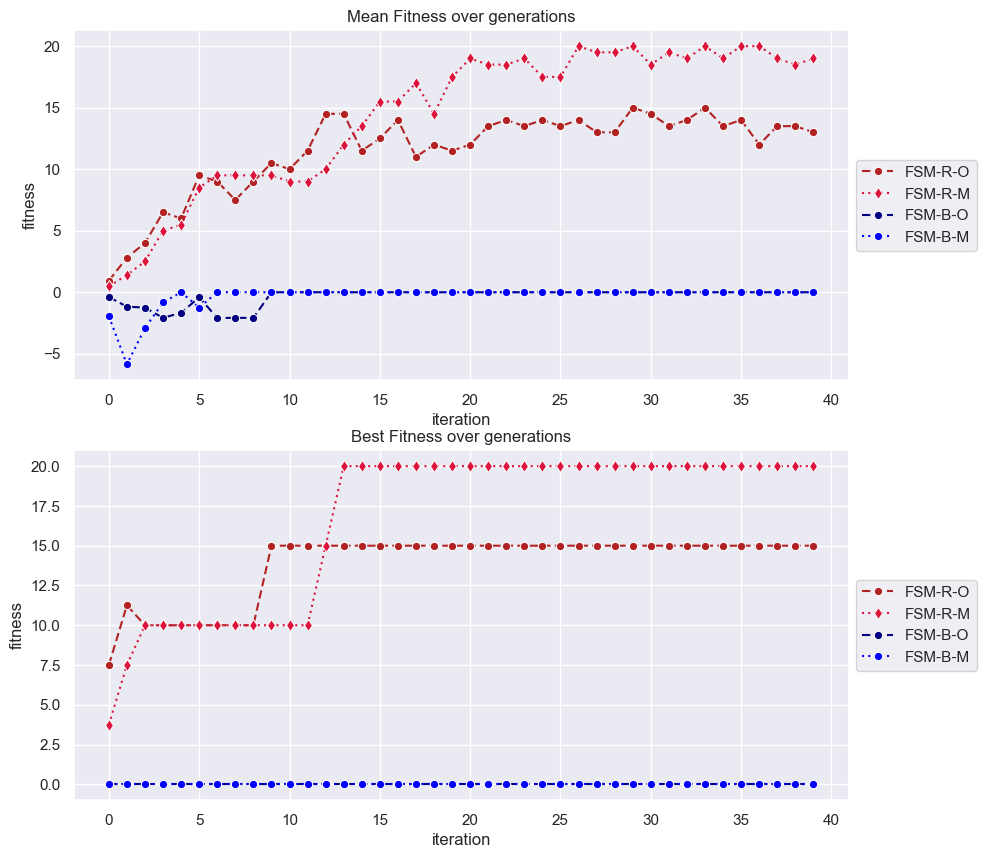}
  \caption{ES (FSM Agent)}
  \label{fig:es_n_controllers}
\end{subfigure}
\begin{subfigure}[b]{0.49\textwidth}
  \centering
  \includegraphics[width=\textwidth]{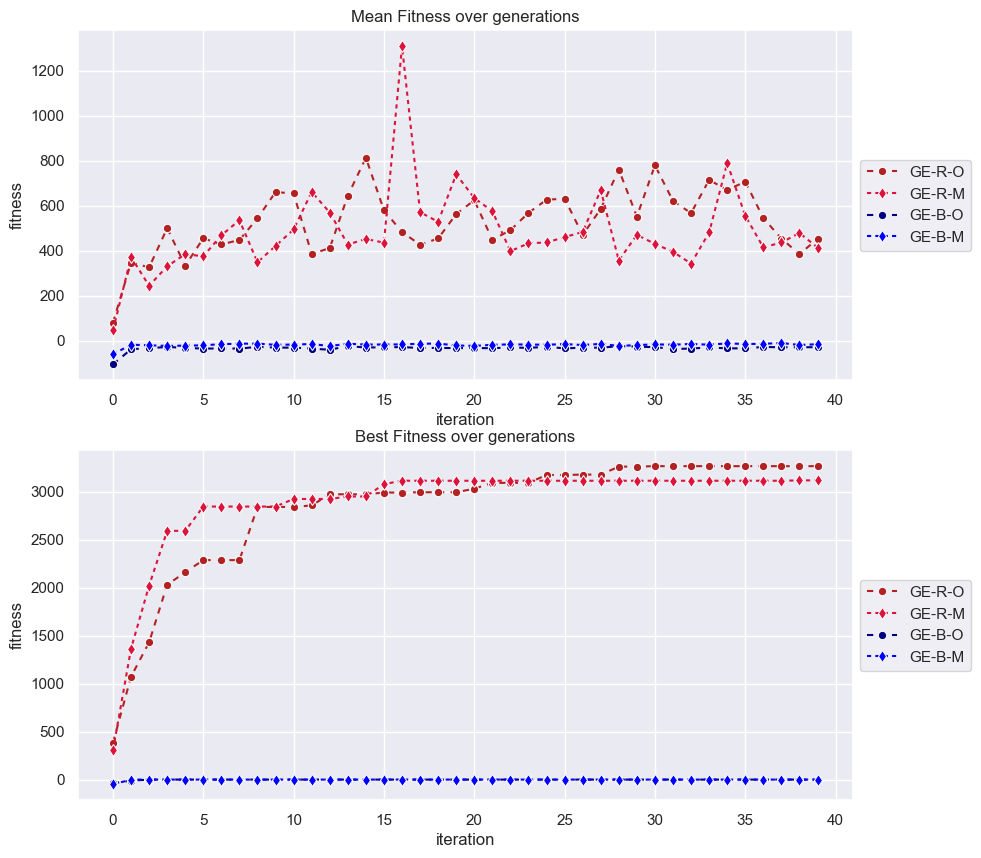}
  \caption{GE (Grammar agent)}
  \label{fig:ge_n_controllers}
\end{subfigure}

\caption{Number of controllers per team. X-axis it the
  iteration. Y-axis agent fitness~(reward). One subplot shows best
  fitness the other mean population fitness. Lines show number of
  controllers per team, (O)ne or (M)any}
  \label{fig:n_controllers}
\end{figure*}

Figure~\ref{fig:n_controllers} shows the evolution of rewards for
different number of evolved controllers per team. For the FSM
representation in Figure~\ref{fig:es_n_controllers} we observe that
multiple controllers improve the training fitness for both red and
blue agents. For the code rule representation in
Figure~\ref{fig:ge_n_controllers} there is no notable difference in
mean population fitness, and maybe slightly better for the red agent
to use one controller per team. 

\subsubsection{Grammar complexity}

\begin{figure*}[!h]
  \centering
    \captionsetup{font=footnotesize}
\begin{subfigure}[b]{0.49\textwidth}
  \centering
  \includegraphics[width=\textwidth]{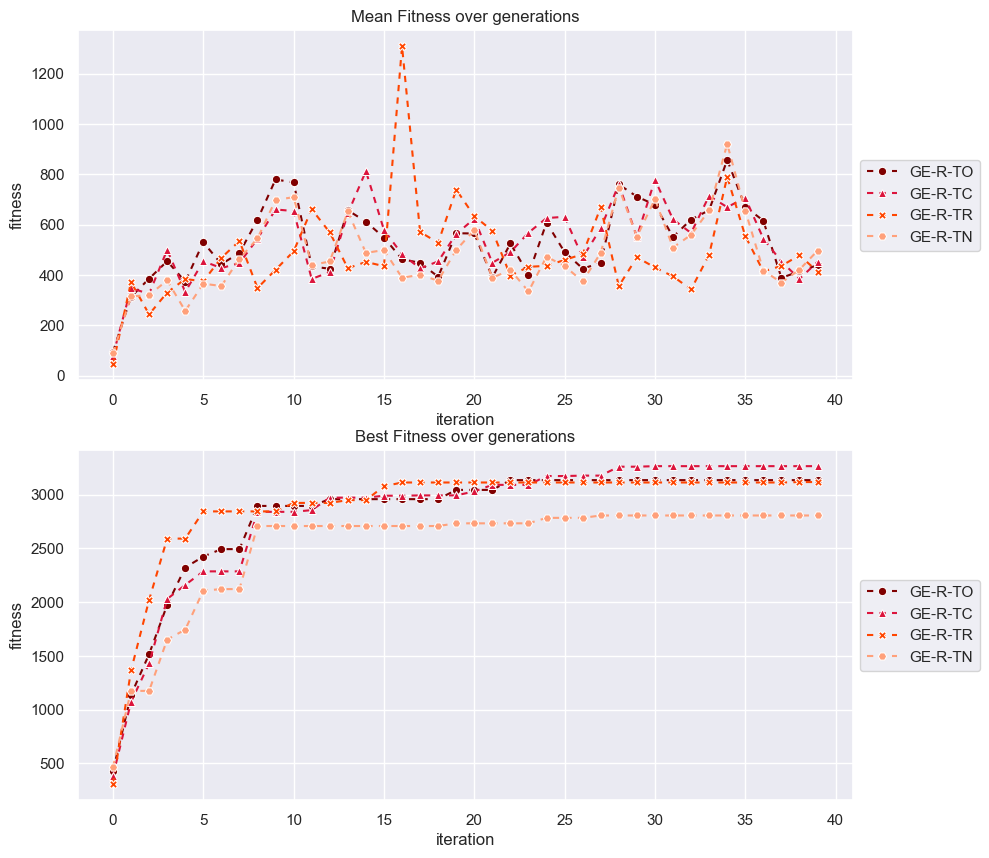}
  \caption{GE-Red}
  \label{fig:ge_target_selection_red}
\end{subfigure}
\begin{subfigure}[b]{0.49\textwidth}
  \centering
  \includegraphics[width=\textwidth]{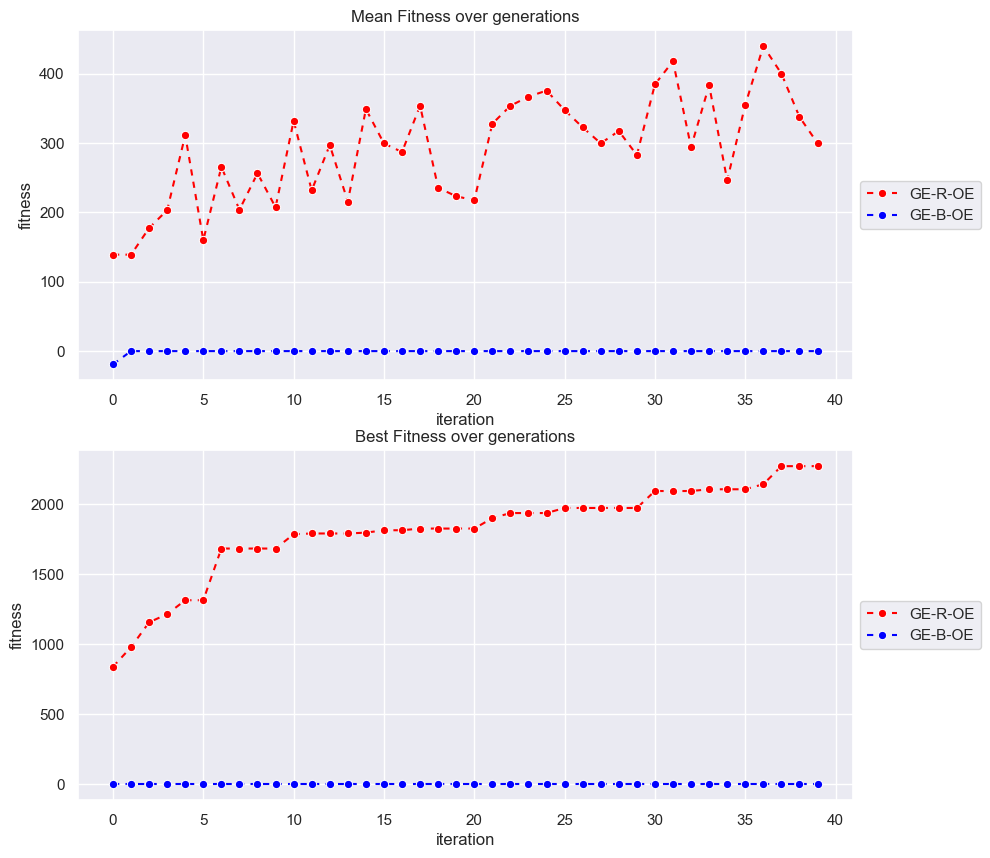}
  \caption{GE Extended Observation Grammar}
  \label{fig:ge_extended_grammar}
\end{subfigure}

\caption{Target selection. X-axis it the iteration. Y-axis
  agent fitness~(reward). One subplot shows best fitness the other
  mean population fitness. Lines show the grammar}
  \label{fig:target_selection}
\end{figure*}

Figure~\ref{fig:ge_target_selection_red} shows the evolution of rewards for
different target selections in the grammar. The red team could have a
difference with the grammar that chooses the newest
target~(GE-R-TN). Note, for the blue team
see~\ref{sec:target_selection_grammar_app}.
Figure~\ref{fig:ge_extended_grammar} shows the evolution of rewards
for extending the number of observation production choices and
functions in the grammar.  Both best and mean fitness improve with
more available observations when compared with
Figure~\ref{fig:ge_c_results}. Thus, we see the impact of providing
the right functions to the agent even though the grammar size
increases.

\subsubsection{LLM-supported Mutation} \label{sec:llm-mu-results}

Figure~\ref{fig:ge_c_results} and \ref{fig:ge_llm_c_results} show GE
and GE-C runs with and without the use of a mutation operator
supported by an LLM (gpt-3.5). We see that the operator makes a
difference. 

\begin{figure}[!h]
  \centering
    \captionsetup{font=footnotesize}
  \includegraphics[width=0.2049\textwidth]{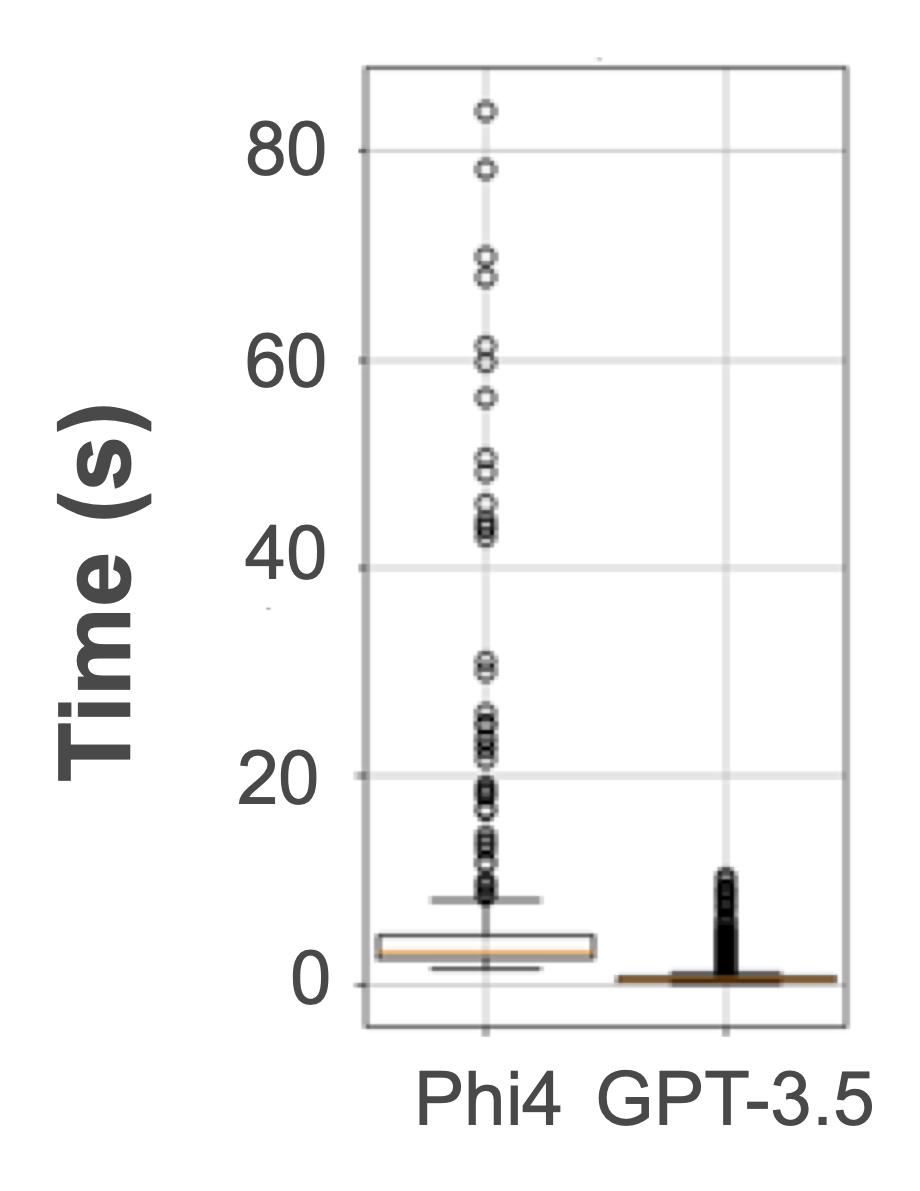}
  \caption{Timing results for GPT 3.5 and Phi4 when used in GE-LLM. }
  \label{fig:timingLLMs}
\end{figure}

Table~\ref{tab:llm_stats} presents the quality of LLM responses and
quantities of the LLM-models' response tokens. We observe longer
responses from GPT-3.5, but with a lower rate of valid
responses. Figure~\ref{fig:timingLLMs} shows how long the LLMs took to
respond to queries. We note that Phi-4 is slower then GPT-3.5. This
could be due to the local installation and hardware of our Phi-4
instance compared to the opaque GPT-.3.5 tier and API
performance~\cite{hemberg2024evolving}.
\begin{table}[]
\caption{Validity of LLM-supported Mutations, Number of Tokens in LLM Responses.}
\label{tab:llm_stats}
\begin{tabular}{l|cc|ccc}
 & \multicolumn{2}{c}{\textbf{Mutations}}       & \multicolumn{3}{|c}{\textbf{Response Tokens}}                         \\ \cline{2-6}
 & \textbf{Valid} & \textbf{Invalid} & \textbf{Min.} & \textbf{Max.} & \textbf{Mean} \\ \hline
\multicolumn{1}{l|}{gpt-3.5-turbo} & \multicolumn{1}{c}{4201 (92\%)} & 351 (8\%) & \multicolumn{1}{c|}{8}  & \multicolumn{1}{c}{863} & 56.6 \\ \hline
\multicolumn{1}{l|}{Phi4}          & \multicolumn{1}{c}{263 (99\%)}  & 3 (1\%)   & \multicolumn{1}{c|}{30} & \multicolumn{1}{c}{513} & 103 
\end{tabular}
\end{table}

The prompts are key to LLM performance. We progressively refined the
prompts to the LLMs.  Our naive, initial prompt simply inserted code
in the prompt and asked for mutated code. LLM responses were very
poor, typically inserting functions that did not exist.  We improved
by including the grammar in the prompt.  This helped the LLM generate
code with better syntax. Mutations were however minor e.g. tweaks of
variables but not changes in the code structure by adding to it
hierarchically or with insertions. To increase the variation, we added
instructions to expand the code and responses made structural changes
such as adding an if statement or adding a condition. The code
remained, however, incoherent in terms of doing something
sensible. When we added info to the prompt saying that the LLM should
consider itself a cyber security expert, we finally started to obtain
high quality mutations with reasonable semantic behavior.

\subsection{Discussion}
\label{sec:discussion}

We consistently observe that the fitness~(reward) from the two-sided
learning of dynamically evolved solutions to be lower than the
one-sided learning of statically evolved solutions. The
representations and algorithms all have different properties and
biases. We observed key properties of representation size, search
space size and domain knowledge. The fixed size representations and
mappings of ES and GA are often suitable for fixed size problems,
e.g. Finite State Machines. In addition, the ES search space size is larger
than the GA search space. The GE representations are
variable length which can be more suitable to variable length
representations like code. The variable length grammar search space of
code depends on the grammar size. This is often larger than the GA
search space. Finally, the fixed mapping of GA and ES can require less
domain knowledge than the grammar used in GE.

\paragraph{Limitations}

There are several limitations. First, no out-of-sample evaluation of
evolved solutions. We also only tried a limited number of
representations. In the \cccFour version we used a limited solution
steps and repetition. In addition, we did not use of messages between
agents. None of the algorithms was explicitly designed for
cooperation, e.g a cooperative coevolutionary algorithm.

\section{Conclusions \& Future Work}
\label{sec:conclusions}

We investigate how we can use evolutionary algorithms to effectively
train autonomous agents in a scenario hosted on CybORG, a simulation
gym designed for cyber security research. Starting from blue training,
we extend the CAGE 4 Challenge Scenario to allow red agents to also be
trained.  This allows us to find that co-evolved agent teams (red vs
blue) attain lower rewards than teams trained against a fixed
adversary. To design for the complexity of the scenario, we progress
through two designs for agent controllers, FSM and grammar. For FSM
the discrete representation works best in our scenario. Furthermore,
we find that the grammar-based agent representation has the highest
training fitness. Moreover, the grammar expressiveness can influence
the performance, and a larger grammar can perform well with the correct
terminals. An LLM based grammar operator is very sensitive to prompt
and LLM-model.

In future work we will investigate further out-of-sample performance
of the trained agents. We will also try more parameter sensitivity
values, as well as grammars. Finally, the transferability of the rule
agents to different scenarios as well as emulation will be
investigated.

\begin{acks}
We acknowledge funding for this work under US Government Contract \#FA8075-18-D-0008.
\end{acks}

\clearpage

\newpage

\balance
\bibliographystyle{ACM-Reference-Format}
\bibliography{references}


\begin{thebibliography}{28}


\ifx \showCODEN    \undefined \def \showCODEN     #1{\unskip}     \fi
\ifx \showDOI      \undefined \def \showDOI       #1{#1}\fi
\ifx \showISBNx    \undefined \def \showISBNx     #1{\unskip}     \fi
\ifx \showISBNxiii \undefined \def \showISBNxiii  #1{\unskip}     \fi
\ifx \showISSN     \undefined \def \showISSN      #1{\unskip}     \fi
\ifx \showLCCN     \undefined \def \showLCCN      #1{\unskip}     \fi
\ifx \shownote     \undefined \def \shownote      #1{#1}          \fi
\ifx \showarticletitle \undefined \def \showarticletitle #1{#1}   \fi
\ifx \showURL      \undefined \def \showURL       {\relax}        \fi
\providecommand\bibfield[2]{#2}
\providecommand\bibinfo[2]{#2}
\providecommand\natexlab[1]{#1}
\providecommand\showeprint[2][]{arXiv:#2}

\bibitem[\protect\citeauthoryear{??}{cag}{2022}]%
        {cage_cyborg_2023}
 \bibinfo{year}{2022}\natexlab{}.
\newblock \bibinfo{title}{Cyber Operations Research Gym}.
\newblock
  \bibinfo{howpublished}{\url{https://github.com/cage-challenge/CybORG}}.
\newblock
\newblock
\shownote{Created by Maxwell Standen, David Bowman, Olivia Naish, Ben Edwards,
  James Drane, Claire Owens, KC Cowan, Wayne Gould, Mitchell Kiely, Son Hoang,
  Toby Richer, Martin Lucas, Richard Van Tassel, Phillip Vu, Natalie Konschnik,
  Joshua Collyer, Calum Fairchild, Thomas Harding}.


\bibitem[\protect\citeauthoryear{Antonio and Coello}{Antonio and
  Coello}{2018}]%
        {Antonio2018}
\bibfield{author}{\bibinfo{person}{L.~M. Antonio} {and}
  \bibinfo{person}{C.~A.~C. Coello}.} \bibinfo{year}{2018}\natexlab{}.
\newblock \showarticletitle{Coevolutionary Multi-objective Evolutionary
  Algorithms: A Survey of the State-of-the-Art}.
\newblock \bibinfo{journal}{\emph{IEEE Transactions on Evolutionary
  Computation}} (\bibinfo{year}{2018}), \bibinfo{pages}{1--16}.
\newblock
\showISSN{1089-778X}
\urldef\tempurl%
\url{https://doi.org/10.1109/TEVC.2017.2767023}
\showDOI{\tempurl}


\bibitem[\protect\citeauthoryear{Back}{Back}{1996}]%
        {back1996evolutionary}
\bibfield{author}{\bibinfo{person}{Thomas Back}.}
  \bibinfo{year}{1996}\natexlab{}.
\newblock \bibinfo{booktitle}{\emph{Evolutionary algorithms in theory and
  practice: evolution strategies, evolutionary programming, genetic
  algorithms}}.
\newblock \bibinfo{publisher}{Oxford university press}.
\newblock


\bibitem[\protect\citeauthoryear{Custode, Migliore~Rambaldi, Roveri, and
  Iacca}{Custode et~al\mbox{.}}{2024}]%
        {custode2024comparing}
\bibfield{author}{\bibinfo{person}{Leonardo~Lucio Custode},
  \bibinfo{person}{Chiara~Camilla Migliore~Rambaldi}, \bibinfo{person}{Marco
  Roveri}, {and} \bibinfo{person}{Giovanni Iacca}.}
  \bibinfo{year}{2024}\natexlab{}.
\newblock \showarticletitle{Comparing large language models and grammatical
  evolution for code generation}. In \bibinfo{booktitle}{\emph{Proceedings of
  the Genetic and Evolutionary Computation Conference Companion}}.
  \bibinfo{pages}{1830--1837}.
\newblock


\bibitem[\protect\citeauthoryear{Ehrlich and Raven}{Ehrlich and Raven}{1964}]%
        {ehrlich1964butterflies}
\bibfield{author}{\bibinfo{person}{Paul~R Ehrlich} {and}
  \bibinfo{person}{Peter~H Raven}.} \bibinfo{year}{1964}\natexlab{}.
\newblock \showarticletitle{Butterflies and plants: a study in coevolution}.
\newblock \bibinfo{journal}{\emph{Evolution}} \bibinfo{volume}{18},
  \bibinfo{number}{4} (\bibinfo{year}{1964}), \bibinfo{pages}{586--608}.
\newblock


\bibitem[\protect\citeauthoryear{Fajardo, Hemberg, Toutouh, O’Reilly, and
  Lehre}{Fajardo et~al\mbox{.}}{2024}]%
        {Fajardo2024ASC}
\bibfield{author}{\bibinfo{person}{Mario Alejandro~Hevia Fajardo},
  \bibinfo{person}{Erik Hemberg}, \bibinfo{person}{Jamal Toutouh},
  \bibinfo{person}{Una-May O’Reilly}, {and} \bibinfo{person}{P. Lehre}.}
  \bibinfo{year}{2024}\natexlab{}.
\newblock \showarticletitle{A Self-adaptive Coevolutionary Algorithm}.
\newblock \bibinfo{journal}{\emph{Proceedings of the Genetic and Evolutionary
  Computation Conference}} (\bibinfo{year}{2024}).
\newblock
\urldef\tempurl%
\url{https://api.semanticscholar.org/CorpusID:271065215}
\showURL{%
\tempurl}


\bibitem[\protect\citeauthoryear{Goel, Ward, Neumann, Neumann, Nguyen, and
  Guo}{Goel et~al\mbox{.}}{2024}]%
        {goel2024hardening}
\bibfield{author}{\bibinfo{person}{Diksha Goel}, \bibinfo{person}{Max Ward},
  \bibinfo{person}{Aneta Neumann}, \bibinfo{person}{Frank Neumann},
  \bibinfo{person}{Hung Nguyen}, {and} \bibinfo{person}{Mingyu Guo}.}
  \bibinfo{year}{2024}\natexlab{}.
\newblock \showarticletitle{Hardening Active Directory Graphs via Evolutionary
  Diversity Optimization based Policies}.
\newblock \bibinfo{journal}{\emph{ACM Transactions on Evolutionary Learning}}
  (\bibinfo{year}{2024}).
\newblock


\bibitem[\protect\citeauthoryear{Goldberg}{Goldberg}{1989}]%
        {Goldberg1989}
\bibfield{author}{\bibinfo{person}{David~E. Goldberg}.}
  \bibinfo{year}{1989}\natexlab{}.
\newblock \bibinfo{booktitle}{\emph{Genetic Algorithms in Search, Optimization
  and Machine Learning} (\bibinfo{edition}{1st} ed.)}.
\newblock \bibinfo{publisher}{Addison-Wesley Longman Publishing Co., Inc.},
  \bibinfo{address}{Boston, MA, USA}.
\newblock
\showISBNx{0201157675}


\bibitem[\protect\citeauthoryear{Group}{Group}{2023}]%
        {CCC4}
\bibfield{author}{\bibinfo{person}{TTCP CAGE~Working Group}.}
  \bibinfo{year}{2023}\natexlab{}.
\newblock \bibinfo{title}{TTCP CAGE Challenge 4}.
\newblock
  \bibinfo{howpublished}{\url{https://github.com/cage-challenge/cage-challenge-4}}.
\newblock


\bibitem[\protect\citeauthoryear{Hammar, Dhir, and Stadler}{Hammar
  et~al\mbox{.}}{2024}]%
        {hammar2024optimal}
\bibfield{author}{\bibinfo{person}{Kim Hammar}, \bibinfo{person}{Neil Dhir},
  {and} \bibinfo{person}{Rolf Stadler}.} \bibinfo{year}{2024}\natexlab{}.
\newblock \showarticletitle{Optimal Defender Strategies for CAGE-2 using Causal
  Modeling and Tree Search}.
\newblock \bibinfo{journal}{\emph{arXiv preprint arXiv:2407.11070}}
  (\bibinfo{year}{2024}).
\newblock


\bibitem[\protect\citeauthoryear{Harris and Tauritz}{Harris and
  Tauritz}{2021}]%
        {Harris2021CompetitiveCF}
\bibfield{author}{\bibinfo{person}{Sean~N. Harris} {and}
  \bibinfo{person}{Daniel~R. Tauritz}.} \bibinfo{year}{2021}\natexlab{}.
\newblock \showarticletitle{Competitive coevolution for defense and security:
  Elo-based similar-strength opponent sampling}.
\newblock \bibinfo{journal}{\emph{Proceedings of the Genetic and Evolutionary
  Computation Conference Companion}} (\bibinfo{year}{2021}).
\newblock
\urldef\tempurl%
\url{https://api.semanticscholar.org/CorpusID:235770343}
\showURL{%
\tempurl}


\bibitem[\protect\citeauthoryear{Heckel}{Heckel}{2023}]%
        {Heckel2023NeuroevolutionFA}
\bibfield{author}{\bibinfo{person}{Kade Heckel}.}
  \bibinfo{year}{2023}\natexlab{}.
\newblock \showarticletitle{Neuroevolution for Autonomous Cyber Defense}.
\newblock \bibinfo{journal}{\emph{Proceedings of the Companion Conference on
  Genetic and Evolutionary Computation}} (\bibinfo{year}{2023}).
\newblock
\urldef\tempurl%
\url{https://api.semanticscholar.org/CorpusID:260119489}
\showURL{%
\tempurl}


\bibitem[\protect\citeauthoryear{Hemberg, Moskal, and O’Reilly}{Hemberg
  et~al\mbox{.}}{2024}]%
        {hemberg2024evolving}
\bibfield{author}{\bibinfo{person}{Erik Hemberg}, \bibinfo{person}{Stephen
  Moskal}, {and} \bibinfo{person}{Una-May O’Reilly}.}
  \bibinfo{year}{2024}\natexlab{}.
\newblock \showarticletitle{Evolving code with a large language model}.
\newblock \bibinfo{journal}{\emph{Genetic Programming and Evolvable Machines}}
  \bibinfo{volume}{25}, \bibinfo{number}{2} (\bibinfo{year}{2024}),
  \bibinfo{pages}{21}.
\newblock


\bibitem[\protect\citeauthoryear{Hemberg, Rosen, Warner, Wijesinghe, and
  O’Reilly}{Hemberg et~al\mbox{.}}{2016}]%
        {hemberg2016detecting}
\bibfield{author}{\bibinfo{person}{Erik Hemberg}, \bibinfo{person}{Jacob
  Rosen}, \bibinfo{person}{Geoff Warner}, \bibinfo{person}{Sanith Wijesinghe},
  {and} \bibinfo{person}{Una-May O’Reilly}.} \bibinfo{year}{2016}\natexlab{}.
\newblock \showarticletitle{Detecting tax evasion: a co-evolutionary approach}.
\newblock \bibinfo{journal}{\emph{Artificial Intelligence and Law}}
  \bibinfo{volume}{24} (\bibinfo{year}{2016}), \bibinfo{pages}{149--182}.
\newblock


\bibitem[\protect\citeauthoryear{Kelly and Heywood}{Kelly and Heywood}{2017}]%
        {Kelly2017EmergentTG}
\bibfield{author}{\bibinfo{person}{Stephen Kelly} {and}
  \bibinfo{person}{Malcolm~I. Heywood}.} \bibinfo{year}{2017}\natexlab{}.
\newblock \showarticletitle{Emergent Tangled Graph Representations for Atari
  Game Playing Agents}. In \bibinfo{booktitle}{\emph{European Conference on
  Genetic Programming}}.
\newblock
\urldef\tempurl%
\url{https://api.semanticscholar.org/CorpusID:26568610}
\showURL{%
\tempurl}


\bibitem[\protect\citeauthoryear{Krawiec and Heywood}{Krawiec and
  Heywood}{2016}]%
        {krawiec2016solving}
\bibfield{author}{\bibinfo{person}{Krzysztof Krawiec} {and}
  \bibinfo{person}{Malcolm Heywood}.} \bibinfo{year}{2016}\natexlab{}.
\newblock \showarticletitle{Solving Complex Problems with Coevolutionary
  Algorithms}. In \bibinfo{booktitle}{\emph{Proceedings of the 2016 on Genetic
  and Evolutionary Computation Conference Companion}}. ACM,
  \bibinfo{pages}{687--713}.
\newblock


\bibitem[\protect\citeauthoryear{Maliukov, Weiss, Margalit, and
  Elyasaf}{Maliukov et~al\mbox{.}}{2024}]%
        {Maliukov2024EvolvingAC}
\bibfield{author}{\bibinfo{person}{Irina Maliukov}, \bibinfo{person}{Gera
  Weiss}, \bibinfo{person}{Oded Margalit}, {and} \bibinfo{person}{Achiya
  Elyasaf}.} \bibinfo{year}{2024}\natexlab{}.
\newblock \showarticletitle{Evolving Assembly Code in an Adversarial
  Environment}.
\newblock \bibinfo{journal}{\emph{ArXiv}}  \bibinfo{volume}{abs/2403.19489}
  (\bibinfo{year}{2024}).
\newblock
\urldef\tempurl%
\url{https://api.semanticscholar.org/CorpusID:268732824}
\showURL{%
\tempurl}


\bibitem[\protect\citeauthoryear{Mitchell}{Mitchell}{2006}]%
        {mitchell2006coevolutionary}
\bibfield{author}{\bibinfo{person}{Melanie Mitchell}.}
  \bibinfo{year}{2006}\natexlab{}.
\newblock \showarticletitle{Coevolutionary learning with spatially distributed
  populations}.
\newblock \bibinfo{journal}{\emph{Computational intelligence: principles and
  practice}}  \bibinfo{volume}{400} (\bibinfo{year}{2006}).
\newblock


\bibitem[\protect\citeauthoryear{O'Neill and Ryan}{O'Neill and Ryan}{2001}]%
        {o2001grammatical}
\bibfield{author}{\bibinfo{person}{Michael O'Neill} {and}
  \bibinfo{person}{Conor Ryan}.} \bibinfo{year}{2001}\natexlab{}.
\newblock \showarticletitle{Grammatical evolution}.
\newblock \bibinfo{journal}{\emph{IEEE Transactions on Evolutionary
  Computation}} \bibinfo{volume}{5}, \bibinfo{number}{4}
  (\bibinfo{year}{2001}), \bibinfo{pages}{349--358}.
\newblock


\bibitem[\protect\citeauthoryear{O’Reilly, Toutouh, Pertierra, Sanchez,
  Garcia, Luogo, Kelly, and Hemberg}{O’Reilly et~al\mbox{.}}{2020}]%
        {o2020adversarial}
\bibfield{author}{\bibinfo{person}{Una-May O’Reilly}, \bibinfo{person}{Jamal
  Toutouh}, \bibinfo{person}{Marcos Pertierra}, \bibinfo{person}{Daniel~Prado
  Sanchez}, \bibinfo{person}{Dennis Garcia}, \bibinfo{person}{Anthony~Erb
  Luogo}, \bibinfo{person}{Jonathan Kelly}, {and} \bibinfo{person}{Erik
  Hemberg}.} \bibinfo{year}{2020}\natexlab{}.
\newblock \showarticletitle{Adversarial genetic programming for cyber security:
  A rising application domain where GP matters}.
\newblock \bibinfo{journal}{\emph{Genetic Programming and Evolvable Machines}}
  \bibinfo{volume}{21} (\bibinfo{year}{2020}), \bibinfo{pages}{219--250}.
\newblock


\bibitem[\protect\citeauthoryear{Popovici, Bucci, Wiegand, and
  De~Jong}{Popovici et~al\mbox{.}}{2012}]%
        {popovici2012}
\bibfield{author}{\bibinfo{person}{Elena Popovici}, \bibinfo{person}{Anthony
  Bucci}, \bibinfo{person}{R.~Paul Wiegand}, {and} \bibinfo{person}{Edwin~D.
  De~Jong}.} \bibinfo{year}{2012}\natexlab{}.
\newblock \bibinfo{booktitle}{\emph{Coevolutionary Principles}}.
\newblock \bibinfo{publisher}{Springer Berlin Heidelberg},
  \bibinfo{address}{Berlin, Heidelberg}, \bibinfo{pages}{987--1033}.
\newblock


\bibitem[\protect\citeauthoryear{Rosin and Belew}{Rosin and Belew}{1997}]%
        {rosin1997new}
\bibfield{author}{\bibinfo{person}{Christopher~D Rosin} {and}
  \bibinfo{person}{Richard~K Belew}.} \bibinfo{year}{1997}\natexlab{}.
\newblock \showarticletitle{New methods for competitive coevolution}.
\newblock \bibinfo{journal}{\emph{Evolutionary Computation}}
  \bibinfo{volume}{5}, \bibinfo{number}{1} (\bibinfo{year}{1997}),
  \bibinfo{pages}{1--29}.
\newblock


\bibitem[\protect\citeauthoryear{Rush, Tauritz, and Kent}{Rush
  et~al\mbox{.}}{2015}]%
        {Rush2015CoevolutionaryAN}
\bibfield{author}{\bibinfo{person}{George Rush}, \bibinfo{person}{Daniel~R.
  Tauritz}, {and} \bibinfo{person}{Alexander~D. Kent}.}
  \bibinfo{year}{2015}\natexlab{}.
\newblock \showarticletitle{Coevolutionary Agent-based Network Defense
  Lightweight Event System (CANDLES)}.
\newblock \bibinfo{journal}{\emph{Proceedings of the Companion Publication of
  the 2015 Annual Conference on Genetic and Evolutionary Computation}}
  (\bibinfo{year}{2015}).
\newblock
\urldef\tempurl%
\url{https://api.semanticscholar.org/CorpusID:15753509}
\showURL{%
\tempurl}


\bibitem[\protect\citeauthoryear{Shashkov, Hemberg, Tulla, and
  O’Reilly}{Shashkov et~al\mbox{.}}{2023}]%
        {Shashkov2023AdversarialAF}
\bibfield{author}{\bibinfo{person}{Alexander Shashkov}, \bibinfo{person}{Erik
  Hemberg}, \bibinfo{person}{Miguel Tulla}, {and} \bibinfo{person}{Una-May
  O’Reilly}.} \bibinfo{year}{2023}\natexlab{}.
\newblock \showarticletitle{Adversarial agent-learning for cybersecurity: a
  comparison of algorithms}.
\newblock \bibinfo{journal}{\emph{The Knowledge Engineering Review}}
  \bibinfo{volume}{38} (\bibinfo{year}{2023}).
\newblock
\urldef\tempurl%
\url{https://api.semanticscholar.org/CorpusID:257354029}
\showURL{%
\tempurl}


\bibitem[\protect\citeauthoryear{Sims}{Sims}{1994}]%
        {sims1994evolving}
\bibfield{author}{\bibinfo{person}{Karl Sims}.}
  \bibinfo{year}{1994}\natexlab{}.
\newblock \showarticletitle{Evolving 3D morphology and behavior by
  competition}.
\newblock \bibinfo{journal}{\emph{Artificial life}} \bibinfo{volume}{1},
  \bibinfo{number}{4} (\bibinfo{year}{1994}), \bibinfo{pages}{353--372}.
\newblock


\bibitem[\protect\citeauthoryear{Smith, Zincir-Heywood, Heywood, and
  Jacobs}{Smith et~al\mbox{.}}{2016}]%
        {Smith2016InitiatingAM}
\bibfield{author}{\bibinfo{person}{Robert~J. Smith}, \bibinfo{person}{Ayse~Nur
  Zincir-Heywood}, \bibinfo{person}{Malcolm~I. Heywood}, {and}
  \bibinfo{person}{John~T. Jacobs}.} \bibinfo{year}{2016}\natexlab{}.
\newblock \showarticletitle{Initiating a Moving Target Network Defense with a
  Real-time Neuro-evolutionary Detector}.
\newblock \bibinfo{journal}{\emph{Proceedings of the 2016 on Genetic and
  Evolutionary Computation Conference Companion}} (\bibinfo{year}{2016}).
\newblock
\urldef\tempurl%
\url{https://api.semanticscholar.org/CorpusID:16359957}
\showURL{%
\tempurl}


\bibitem[\protect\citeauthoryear{Standen, Lucas, Bowman, Richer, Kim, and
  Marriott}{Standen et~al\mbox{.}}{2021}]%
        {Standen2021CybORGAG}
\bibfield{author}{\bibinfo{person}{Maxwell Standen}, \bibinfo{person}{Martin
  Lucas}, \bibinfo{person}{David Bowman}, \bibinfo{person}{Toby~J. Richer},
  \bibinfo{person}{Junae Kim}, {and} \bibinfo{person}{Damian~A. Marriott}.}
  \bibinfo{year}{2021}\natexlab{}.
\newblock \showarticletitle{CybORG: A Gym for the Development of Autonomous
  Cyber Agents}.
\newblock \bibinfo{journal}{\emph{ArXiv}}  \bibinfo{volume}{abs/2108.09118}
  (\bibinfo{year}{2021}).
\newblock
\urldef\tempurl%
\url{https://api.semanticscholar.org/CorpusID:237259783}
\showURL{%
\tempurl}


\bibitem[\protect\citeauthoryear{Wei, Bierbrauer, Nack, Pavlik, and
  Bastian}{Wei et~al\mbox{.}}{2024}]%
        {Wei2024OfflineRL}
\bibfield{author}{\bibinfo{person}{Alexander Wei}, \bibinfo{person}{David~A.
  Bierbrauer}, \bibinfo{person}{Emily~A. Nack}, \bibinfo{person}{John Pavlik},
  {and} \bibinfo{person}{Nathaniel Bastian}.} \bibinfo{year}{2024}\natexlab{}.
\newblock \showarticletitle{Offline Reinforcement Learning for Autonomous Cyber
  Defense Agents}.
\newblock \bibinfo{journal}{\emph{2024 Winter Simulation Conference (WSC)}}
  (\bibinfo{year}{2024}), \bibinfo{pages}{1978--1989}.
\newblock
\urldef\tempurl%
\url{https://api.semanticscholar.org/CorpusID:275773130}
\showURL{%
\tempurl}


\end{thebibliography}

\clearpage

\appendix

\section{CybORG CAGE Challenge 4}
\label{appendix:CybORG}
The CybORG simulated environment comprises agents
interacting with a scenario that is modeled with finite states
executed in discrete steps. An action has pre-conditions and
post-conditions~(effects). A  state has specific information and events,
e.g. creation and deletion of individual files, or the making or
breaking of network connections.  Once the automated agent is trained,
it can be tested in an emulator, which comprises of AWS virtual
machines to create a high fidelity environment for the agent.

A CybORG scenario defines the environment of agent competition. It
 defines what agents exist, available actions, initial state,
and reward calculation. It also defines the configuration of each host
and the network connections. Actions are defined in the scenario, and
are based cybersecurity professional actions.

After an agent takes an action an observation that describes the new
state of the system as observed by that agent. Each agent is only pro-
vided data for what it could conceivably observe, depending on the
reconnaissance tools available to that agent. The observation from
CybORG is a dictionary. The key ‘success’ indicates if the previous
action was successful, unsuccessful or ‘unknown’. The other keys are
host ids that contains further observations from that host in the
categories Interface, Session, User, System and Process.

The mission in the scenario has three different phases Phase 1, Phase
2A and Phase 2B. Each phase has different network connectivity. During
Phase 1 all tasks operating in each zone have low priority . During
Phase 2A only tasks operating in zones within Deployed Network A have
high priority , while all other tasks have low priority. During Phase
2B only tasks in Deployed Network B have high priority ,
all other tasks have low priority.

\label{sec:cyborg-cage-chall}
\begin{table}[h]
  \centering
  \scriptsize
\caption{Rewards}
  \label{tab:rewards_phase_1}
  \begin{tabular}{l|lll}
\textbf{Zone} &	\textbf{Local Work Fails} & \textbf{Access Service Fails} & \textbf{Red impact/access} \\
\hline
\multicolumn{4}{c}{\textbf{Phase 1}} \\
\hline
HQ Network &	-1 & 	-1 & 	-3\\
Contractor Network &	0 &	-5 & 	-5\\
Restricted Zone A &	-1 &	-3 &	-1\\
Operational Zone A &	-1 &	-1 &	-1\\
Restricted Zone B &	-1 &	-3 &	-1\\
Operational Zone B &	-1 &	-1 &	-1\\
Internet &	0 &	0 &	0 \\
\hline
\multicolumn{4}{c}{\textbf{Phase 2A}} \\
\hline
HQ Network &	-1 & 	-1 & 	-3\\
Contractor Network &	0 &	0 & 	0\\
Restricted Zone A &	-2 &	-1 &	-3\\
Operational Zone A &	-10 &	0 &	-10\\
Restricted Zone B &	-1 &	-1 &	-1\\
Operational Zone B &	-1 &	-1 &	-1\\
Internet &	0 &	0 &	0 \\
\hline
\multicolumn{4}{c}{\textbf{Phase 2B}} \\
\hline
HQ Network &	-1 & 	-1 & 	-3\\
Contractor Network &	0 &	0 & 	0\\
Restricted Zone A &	-1 &	-3 &	-3\\
Operational Zone A &	-1 &	-1 &	-1\\
Restricted Zone B &	-2 &	-3 &	-3\\
Operational Zone B &	-10 &	0 &	-10\\
Internet &	0 &	0 &	0 \\
  \end{tabular}
  \end{table}

\subsection{Code Grammars}
\label{sec:code-grammars}

\paragraph{Blue}
Example grammar for conditional target selection
\begin{small}
\begin{verbatim}
sections: "def select_action_and_target(observation, name):
             "#Select action"
             statements
             "#Select target"
             th_statements
             "return action, target_heuristic
          
statements: statement 
          | statement
            statements
statement: "if" conditions ":
               statement
         |  "action =" actions

th_statements: th_statement 
          | th_statement
            th_statements
th_statement: "if" conditions ":''
                 th_statement 
         |  "target_heuristic =" target_heuristic

target_heuristic: "random_target"
                | "first_target"
                | "last_target"

conditions: operator "and" operator
          | operator "or" operator
          | operator
operator: observations operand constant
         | success "== observation['success']"
operand: ">" 
        | "<" 
        | "=="
success: "TRUE" 
       | "FALSE" 
       | "UNKNOWN"
observations: "connections(observation)"
            | "files_user(observation)"
            | "files_root(observation)"
constant: "0"
        | "1"
        | "2"
actions:  "AllowTrafficZone"
       | "BlockTrafficZone"
       | "Monitor"
       | "Analyse"
       | "Restore"
       | "Remove"
       | "DeployDecoy"
       | "Sleep"
\end{verbatim}
\end{small}

\paragraph{Red}
Example grammar for conditional target selection

\begin{small}
\begin{verbatim}
sections: "def select_action_and_target(observation, name):
             "#Select action"
             statements
             "#Select target"
             th_statements
             "return action, target_heuristic
          
statements: statement 
          | statement
            statements
statement: "if" conditions ":
               statement
         |  "action =" actions

th_statements: th_statement 
          | th_statement
            th_statements
th_statement: "if" conditions ":''
                 th_statement 
         |  "target_heuristic =" target_heuristic

target_heuristic: "random_target"
                | "first_target"
                | "last_target"

conditions: operator "and" operator
          | operator "or" operator
          | operator
operator: observations operand constant
         | success "== observation['success']"
operand: ">" 
        | "<" 
        | "=="
success: "TRUE" 
       | "FALSE" 
       | "UNKNOWN"
observations: "connections(observation)"
            | "files_user(observation)"
            | "files_root(observation)"
            | "n_servers(observation)" 
            | "root_access_levels(observation, name)"
constant: "0"
        | "1"
        | "2"
actions: "DiscoverRemoteSystems" 
       | "AggressiveServiceDiscovery" 
       | "StealthServiceDiscovery" 
       | "ExploitRemoteService" 
       | "PrivilegeEscalate" 
       | "DegradeServices" 
       | "DiscoverDeception" 
       | "Impact" 
       | "Withdraw" 
       | "Sleep"
\end{verbatim}
\end{small}

\subsection{Finite State Machines}
\label{sec:finite-state-mach}

\paragraph{Red}

K: Known, KD: Known with Discovery, S: Service, SD: Service with Discovery, U: User, UD: Usesr with Discovery, R: Root, Root with discovery
\begin{verbatim}
"K": [0.5, 0.25, 0.25, None, None, None, None, None, None],
"KD": [None, 0.5, 0.5, None, None, None, None, None, None],
"S": [0.25, None, None, 0.25, 0.5, None, None, None, None],
"SD": [None, None, None, 0.25, 0.75, None, None, None, None],
"U": [0.5, None, None, None, None, 0.5, None, None, 0.0],
"UD": [None, None, None, None, None, 1.0, None, None, 0.0],
"R": [0.5, None, None, None, None, None, 0.25, 0.25, 0.0],
"RD": [None, None, None, None, None, None, 0.5, 0.5, 0.0],
\end{verbatim}

State Priorities: K, KD, S, SD, U, UD, R, RD

\paragraph{Blue}

CN: network is clean
SN: network is suspicious
DN: network is compromised
CM: machine is clean
SM: machine is suspicious
DM: machine is compromised

\begin{verbatim}
"CN": [0.2, 0.2, 0.2, None, None, 0.2, 0.2],
"SN": [None, 0.2, 0.2, 0.2, 0.2, 0.2, None],
"DN": [None, 0.2, 0.2, 0.2, 0.2, 0.2, None],
"CM": [None, None, 0.5, None, None, 0.25, 0.25],
"SM": [None, 0.2, 0.2, 0.2, 0.2, 0.2, None],
"DM": [None, None, None, 0.5, 0.3, 0.2, None],
\end{verbatim}

State Priorities: CN, SN, DN, CM, SM, DM

\section{Results}

\subsection{Target selection grammars}
\label{sec:target_selection_grammar_app}
\begin{figure}[!h]
  \centering
    \captionsetup{font=footnotesize}
\begin{subfigure}[b]{0.49\textwidth}
  \centering
  \includegraphics[width=\textwidth]{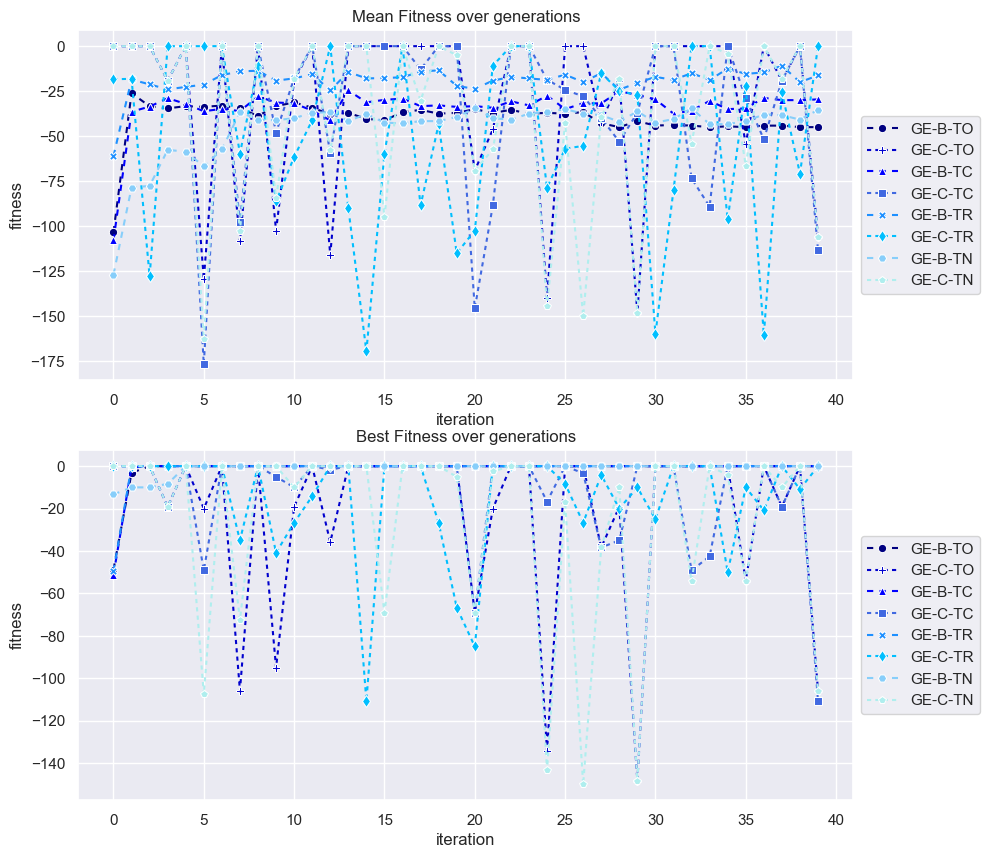}
  \caption{GE-Red}
  \label{fig:ge_target_selection_blue}
\end{subfigure}

\caption{Target selection. X-axis it the generation. Y-axis
  agent fitness~(reward). One subplot shows best fitness the other
  mean population fitness. Lines show the grammar}
  \label{fig:target_selection_blue}
\end{figure}

\end{document}